# The Impact of Battery Cell Configuration on Electric Vehicle Performance: An XGBoost-Based Classification with SHAP Interpretability

Santanam Wishal[1], Louis Filiepe Tio Jansel[2], Matthew Abednego Inkiriwang[3], Jason Sebastian[4]

[1]*Asia Cyber University*
[1]santawishal17@gmail.com

[2]*Bunda Mulia University*
[2]louisftj2@gmail.com

[3]*Pelita Harapan University*
[3]matthewabednego63@gmail.com

[4]*Bina Nusantara University*
[4]jason.js700@gmail.com

*Abstract* — **As the electric vehicle (EV) market continues to prioritize dynamic performance and rapid charging, battery configuration has rapidly evolved. Despite this, current literature has often overlooked the complex, non-linear relationship between battery configuration and electric vehicle performance. To address this gap, this study proposes a machine learning framework which categorizes the EV acceleration performance into High (<= 4.0 seconds), Mid (4.0 - 7.0 seconds), and Low (> 7.0 seconds). Utilizing a preprocessed dataset consisting of 276 EV samples, an Extreme Gradient Boosting (XGBoost) classifier was utilized, achieving 87.5% predictive accuracy, a 0.968 ROC-AUC, and a 0.812 MCC. In order to ensure engineering transparency SHapley Additive exPlanations (SHAP) were employed. Results of analysis shows that an increase in battery cell count initially boosts power delivery, but its mass and complexity diminished performance gains eventually. As such, these findings indicate that battery configuration in EVs must balance system complexity and architectural configuration in order to receive and retain optimal vehicle performance.**

## I. Introduction

### A. The Epoch of Electrification: Global Market Dynamics and the Performance Imperative

The global automotive sector is currently traversing a period of unprecedented disruption, a historical inflection point that marks the decisive transition from the century-long dominance of the Internal Combustion Engine (ICE) to the era of electric mobility. This shift, once driven primarily by niche environmental advocacy and tentative regulatory experiments, has metastasized into a fundamental reorganization of the global industrial economy. As of 2025, the electrification of transport is no longer a prospective scenario but an entrenched reality, characterized by exponential market growth, profound technological maturation, and a radical realignment of competitive hierarchies. The data emanating from the first half of the decade paints a picture of a market that has moved beyond the "early adopter" phase and is rapidly conquering the "early majority," driven by a convergence of strict decarbonization mandates, significant leaps in battery energy density, and a rapidly diversifying consumer base that demands performance parity—if not superiority—over fossil-fuel incumbents [1].

The scale of this adoption is empirically stark. By the first quarter of 2025, global electric vehicle (EV) sales had surged to over 4 million units, registering a robust 35% increase compared to the same period in 2024 [1]. Projections for the full calendar year of 2025 indicate that global sales will exceed 20 million units, a 25% year-on-year expansion that elevates the electric vehicle to a 25% share of the total global automotive market [1]. This statistic—one in every four new cars sold worldwide being electric—represents a critical tipping point in the technology adoption lifecycle. It suggests that the infrastructure and supply chains supporting electrification have achieved a level of resilience and scale capable of supporting mass-market volume, despite lingering macroeconomic uncertainties and the gradual phase-out of purchase subsidies in major Western economies [2].

However, to view this transition merely through the lens of aggregate global volume is to miss the nuanced structural shifts occurring at the regional and manufacturer levels. The "Introduction" of any contemporary analysis on EV performance must account for the heterogeneous nature of this growth, as regional demands dictate the engineering constraints—and thus the battery configurations—of the vehicles being produced.
● **Europe:** The European continent remains the regulatory anvil upon which the EV transition is being forged.

Despite economic headwinds and trade uncertainties, the European EV sales share is forecast to stabilize at approximately 25**%** in 2025, driven largely by the tightening noose of $CO_2$ emission standards [1]. The regulatory framework has created a "comply or pay" environment where Original Equipment Manufacturers (OEMs) prioritize EV volume to offset the emissions of their remaining ICE fleets. The United Kingdom serves as a prime example of policy-driven acceleration; the implementation of the Vehicle Emissions Trading Scheme, which mandated a 22% Zero Emission Vehicle (ZEV) share for 2024, catalyzed a market response that pushed actual sales to nearly 30% [1]. Meanwhile, Norway continues to stand as the global vanguard, with Battery Electric Vehicles (BEVs) accounting for 88% of new car sales, effectively rendering the internal combustion engine obsolete in the new vehicle market and providing a living laboratory for the long-term behavior of EV batteries in cold climates [1].

- **China:** If Europe is the regulatory anvil, China is the industrial forge. The Chinese market has cemented its status not just as the largest consumer of EVs, but as the primary exporter of EV technology. In 2024, China accounted for 40% of global electric car exports, shipping nearly 1.25 million vehicles to markets worldwide [2]. This export dominance is underpinned by a vertically integrated supply chain where Chinese firms control the vast majority of cathode, anode, and cell manufacturing capacity [3]. The domestic market in China is equally dynamic, with New Energy Vehicles (NEVs) set to reach 51.6% of light-vehicle sales in 2025, aiming for a staggering 73% by 2030 [4]. This hyper-competitive environment serves as a crucible for rapid technological iteration, birthing innovations like Cell-to-Pack (CTP) architectures that are central to the performance analysis of this study.
- **The "Leapfrog" Markets:** Perhaps the most significant development of the 2024-2025 period is the emergence of the "EV Leapfrog" in the Global South. Developing economies are bypassing the gradual hybridization phase typical of Western markets and jumping directly to full electrification.
- **Vietnam:** In a remarkable surge, Vietnam's EV sales share doubled to nearly 40% in 2025, a penetration rate that now exceeds that of established markets like the UK and the European Union [5].
- **Indonesia:** Leveraging its massive nickel reserves to build a domestic battery industry, Indonesia has reached a 15% EV market share, overtaking the United States in terms of penetration [5].
- **Latin America:** Brazil and Mexico have both surpassed Japan in EV adoption rates, signaling a shift in the center of gravity for automotive demand [5].

This global diversification of demand fundamentally alters the engineering requirements for electric vehicles. A battery pack designed for the temperate, well-paved roads of Northern Europe may face different thermal and vibrational stresses than one operating in the tropical, congested, and varied topographies of Southeast Asia or Latin America. Consequently, the internal architecture of the battery—specifically the robust configuration of its cells—becomes a critical variable in ensuring reliable performance across these diverse operating domains [2].

### B. The Competitive Landscape and the Redefinition of Performance

The 2025 timeframe also witnessed a historic restructuring of the automotive competitive hierarchy. For the first time, the Chinese automaker BYD overtook Tesla to become the world's largest seller of battery electric vehicles.

- **BYD (2025):** ~2.26 million BEVs delivered, representing a 28% year-on-year growth [6].
- **Tesla (2025):** ~1.64 million BEVs delivered, reflecting a contraction of approximately 9-10% [6].

This "changing of the guard" is not merely a commercial footnote; it is a validation of specific engineering philosophies. BYD's ascendancy is inextricably linked to its proprietary "Blade Battery" technology—a Lithium Iron Phosphate (LFP) based Cell-to-Pack architecture that prioritizes safety and volumetric density [7]. By innovating at the *cell configuration* level, BYD was able to offer vehicles that balanced cost and performance more effectively than competitors relying on traditional modular architectures.

In this hyper-competitive environment, consumer expectations have evolved. The early-market fixation on "Range Anxiety" (the fear of running out of power) is gradually being supplemented by "Performance Anxiety" and "Charging Anxiety." With over 785 EV models available in 2024 and projections hitting 1,000 by 2026, consumers are faced with a paralyzing array of choices [1]. They are increasingly scrutinizing vehicles not just for how far they can go, but for how fast they accelerate, how quickly they charge, and how consistently they deliver power.

Research indicates that while EVs generally compare favorably to ICE vehicles in acceleration, significant variances exist based on battery design. A "High Performance" classification—typically defined by 0-100 km/h acceleration times of under 4 seconds—is no longer the exclusive domain of six-figure supercars; it is becoming a benchmark for premium sedans and SUVs. Achieving this level of performance requires more than just a powerful motor; it requires a battery pack capable of discharging immense current without significant voltage sag or thermal derating. This necessitates a deep understanding of the internal battery configuration—specifically the **Number of Cells** and their electrical topology—which serves as the primary reservoir and pump for the vehicle's kinetic energy.

### C. The Evolution of Battery Architecture: From Passive Modules to Structural Integration

To understand why cell configuration is a determinant of performance, one must examine the rapid evolution of battery packaging technology. The battery pack is the single most expensive and heavy component of an EV, accounting for

30-40% of the total vehicle cost and a significant portion of its curb weight [8]. The drive to optimize this component has moved through three distinct architectural generations.

*1). Generation 1: The Modular Standard*

For the first decade of the modern EV era (approx. 2010-2020), the industry coalesced around a "Cell-Module-Pack" hierarchy. In this approach:
1. **Cells** (the fundamental electrochemical units, packaged as cylinders, pouches, or prismatic cans) are grouped into **Modules**.
2. **Modules** are wired together, monitored by slave Battery Management System (BMS) boards, and encased in protective aluminum or steel housings.
3. **Modules** are then bolted into a master **Pack**, which contains the cooling system, high-voltage busbars, and crash protection [8].

While this modular approach offered advantages in safety (isolating thermal runaway) and serviceability (replacing individual modules), it was inherently inefficient in terms of mass and volume. The module housings, end plates, and interconnects represent "parasitic mass"—passive weight that stores no energy but burdens the vehicle during acceleration [9]. In many modular packs, the "Volumetric Cell-to-Pack Ratio" (the volume of cells divided by the volume of the pack) was often below 40-50%, meaning half the space in the battery was occupied by inactive structural material [9].

*2). Generation 2: Cell-to-Pack (CTP) Technology*

The need to break through the energy density ceiling led to the development of Cell-to-Pack (CTP) technology. Pioneered by companies like CATL and BYD, CTP eliminates the intermediate module level entirely. Cells are integrated directly into the battery pack casing, using the cells themselves as structural members or utilizing advanced adhesives and cooling plates to provide rigidity [9].

TABLE I
Comparative Advantages of Cell-to-Pack (CTP) Technology

| Feature | Impact of CTP vs. Modular | Implication for Performance |
|---|---|---|
| Space Utilization | +15% to +50% usable volume [9]. | Allows for a significantly higher **Number of Cells** in the same wheelbase. |
| Parts Reduction | -40% fewer components [9]. | Reduces "parasitic mass," improving the Power-to-Weight ratio. |
| Energy Density | +10% to +15% (Pack Level) [9]. | Increases range and sustained power output. |
| Manufacturing | 50% increase in production efficiency [10]. | Lowers cost, enabling high-performance batteries in lower-tier segments. |

The transition to CTP is pivotal for this study because it fundamentally alters the **Cell Count** variable. By freeing up 15-50% more space, engineers can pack more active electrochemical material into the vehicle. This is not just about extending range; increasing the number of cells in parallel directly reduces the pack's internal resistance, thereby increasing its maximum power discharge capability [8]. Thus, a vehicle with a CTP battery is likely to have a higher cell count and superior acceleration characteristics compared to a similarly sized vehicle with a modular pack.

*3). Generation 3: Cell-to-Chassis (CTC) and Body (CTB)*

Pushing integration further, Cell-to-Chassis (CTC) and Cell-to-Body (CTB) designs integrate the battery cells directly into the vehicle's frame, effectively turning the battery into the floor of the car. This eliminates the separate battery pack enclosure entirely, further reducing weight and increasing torsional rigidity [11]. While mostly found in newer models (2024-2025), CTC designs represent the ultimate expression of the "mass compounding" benefit—where weight savings beget performance gains—further strengthening the correlation between advanced configuration and acceleration.

*D. The Voltage Revolution: 400V vs. 800V Architectures*

Parallel to the structural evolution of the pack is the electrical evolution of the system voltage. For years, the 400V architecture (typically operating between 300V and 500V) was the industry standard [12]. However, the pursuit of "ultra-fast" charging (350 kW+) and high-performance driving is driving a migration to 800V architectures (operating between 600V and 900V).

This shift is governed by fundamental electrical physics:

$$Power(P) = Voltage(V) \times Current(I)$$
$$Power\ Loss = I^2 \times Resistance(R)$$

To deliver high power—essential for sub-4-second acceleration—a 400V system requires very high current. High current generates significant heat in the cables and battery interconnects (scaling with the square of the current), necessitating heavy copper cabling and robust cooling systems to prevent overheating [13].

By doubling the voltage to 800V, an EV can deliver the same power with half the current, or double the power with the same current. This reduces resistive heating losses significantly, allowing for sustained high-performance driving without thermal derating [12].

Crucially for this study, the voltage architecture dictates the Number of Cells.
A lithium-ion cell has a nominal voltage of approx. 3.7V.
- **400V System:** Requires ~96 cells connected in series (96 x 3.7V ~ 355V).
- **800V System:** Requires ~192 cells connected in series (192 x 3.7V ~ 710V) [14].

Therefore, an 800V vehicle will inherently have a higher series cell count than a 400V vehicle. This makes the "Number of Cells" variable a powerful proxy for the vehicle's underlying electrical architecture. A high cell count in the dataset likely indicates either a large capacity (many parallel cells) or a high-voltage architecture (many series cells), both of which are traits associated with high-performance vehicles [15].

*E. The Physics of Performance: Connecting Cell Count to Acceleration*

The central hypothesis of this report is that the **Number of Cells** is a primary, yet non-linear, driver of vehicle acceleration. This relationship is rooted in the electromechanical principles of the powertrain.

*1). Internal Resistance and Voltage Sag*

Acceleration is limited by the peak power the battery can supply to the inverter. As current is drawn from the battery, voltage sags due to internal resistance (R) according to the formula:

$$V_{terminal} = V_{OCV} - (I_{load} \times R)$$

If the voltage sags too low, the inverter reaches its low-voltage cutoff or power limit, curtailing acceleration. Increasing the number of cells generally lowers the total pack resistance.

- **Parallel Addition:** Adding cells in parallel divides the resistance ($R_{total} = R_{cell} / N_{parallel}$).
- **Series Addition:** Adding cells in series increases voltage, which reduces the current load needed for a given power demand, thereby reducing the voltage drop impact relative to the total voltage [16].

Thus, a higher total cell count enables the pack to sustain higher power output with less voltage sag, directly translating to faster acceleration times [17].

*2). The Problem of Mass: Diminishing Returns*

However, the relationship is not purely additive. Newton's Second Law (F=ma) governs the vehicle's motion.

$$Acceleration\ (A) = \frac{Force\ (F)}{Mass\ (m)}$$

Adding cells increases the Force (F) potential (via power capability), but it strictly increases the Mass (m).

- **Phase 1 (Optimization):** For a small vehicle, adding cells dramatically increases power with a manageable weight penalty. Acceleration improves linearly.
- **Phase 2 (Saturation):** As the pack becomes larger, the weight of the additional cells—plus the heavier chassis, suspension, and brakes needed to carry them—begins to offset the power gains. The acceleration improvement slows down.
- **Phase 3 (Regression):** In extreme cases, adding more mass yields negligible or even negative returns on acceleration, as the vehicle becomes too heavy for the traction limits of the tires or the efficiency limits of the motors [18].

This physical reality creates a **non-linear** relationship between cell count and performance, characterized by distinct "knees" and plateaus. This complexity explains why simple linear statistical models often fail to predict EV performance accurately and why advanced Machine Learning techniques are required.

*F. Methodological Challenges in Existing Literature*

The academic literature on EVs has historically focused on two distinct domains: macro-level market analysis and micro-level battery chemistry. There is a "missing middle" in the analysis of vehicle-level performance classification based on architectural configuration.

*1). Limitations of Traditional Statistical Models*

Many prior studies utilize Linear Regression (LR) to model vehicle range or consumption [19]. While LR is interpretable and computationally cheap, it relies on the assumption of linearity (Y = mX + c). As established in Section 1.5, the physics of EV performance are inherently non-linear. Battery degradation, thermal behavior, and the power-to-weight trade-off all exhibit complex, curved relationships [20].

For example, a study by Mishra et al. used LR for range estimation but found that it was outperformed by Deep Learning models due to its inability to capture anomalies and complex interactions [19]. Similarly, Ozkan et al. found that capturing the uncertainty in driver behavior required non-linear quantile regression [19]. Purely statistical studies often struggle to account for the "tipping points" in battery design, such as the transition from 400V to 800V or the sudden weight savings of CTP technology.

*2). The Rise of Machine Learning (XGBoost)*

To overcome these limitations, the field is increasingly turning to Machine Learning (ML). Extreme Gradient Boosting (XGBoost) has emerged as a "gold standard" for tabular engineering data.

- **Mechanism:** XGBoost is an ensemble technique that builds sequential decision trees. Each tree attempts to correct the errors of the previous one. This allows the model to capture complex, jagged decision boundaries (e.g., distinguishing between a "Mid" and "High" performance car based on a specific combination of weight and cell count).
- **Applications:** Literature shows XGBoost effectively predicting Battery State of Charge (SOC) (Lei, 2024), State of Health (SOH) (Oyucu et al., 2024), and detecting battery faults (Khan et al., 2024) with higher accuracy than Random Forest or Support Vector Machines.
- **Suitability:** XGBoost is particularly robust against the "curse of dimensionality" and handles missing data well, which is critical when dealing with diverse vehicle specification datasets where manufacturers may not report every parameter.

*3). The Interpretability Gap and SHAP*

While XGBoost offers superior predictive performance, it suffers from the "Black Box" problem. In an engineering context, a prediction without a physical explanation is dangerous. Knowing that a car is high-performance is less valuable than knowing why.

To bridge this gap, this study employs SHAP (SHapley Additive exPlanations). Originating from cooperative game theory, SHAP assigns a "payout" (importance value) to each feature for every prediction. This allows us to visualize the exact shape of the relationship—for instance, plotting the SHAP value of "Number of Cells" against the actual cell count to reveal the diminishing returns curve.1 Recent work by Chen et al. (2025) successfully used SHAP to explain battery degradation factors, proving its utility in this domain [21].

*G. Research Gap, Problem Statement, and Contribution*

Despite the rich body of work on battery chemistry and market trends, a specific gap remains:

There is a lack of comprehensive studies that classify Electric Vehicle Acceleration Performance based primarily on Battery Cell Configuration (specifically Cell Count) using interpretable non-linear Machine Learning.

**Most existing studies:**
1. **Focus on Range:** Prioritizing energy consumption over dynamic performance (acceleration/torque).
2. **Use Aggregate Metrics:** Relying on "Battery Capacity (kWh)" rather than the more structurally significant "Number of Cells".
3. **Lack Interpretability:** Using "Black Box" neural networks or standard regression without providing granular engineering insights into *why* certain configurations yield better performance.

**Research Objectives:**
This report aims to fill this void by:
1. **Constructing a robust dataset** of 2025 EV models, specifically isolating the "Number of Cells" as a key architectural feature.
2. **Developing a Multi-Class XGBoost Classifier** to categorize vehicles into High, Mid, and Low acceleration classes.
3. **Applying SHAP Analysis** to deconstruct the model, quantitatively demonstrating the non-linear impact of cell count and its interactions with vehicle weight.

By doing so, this study provides a bridge between the macroscopic market trends of 2025—the push for performance, the shift to CTP/800V—and the microscopic engineering decisions regarding cell configuration. It offers a validated, interpretable framework for understanding how the internal architecture of the battery defines the driving experience of the modern electric vehicle.

## II. Related Works

The pursuit of understanding and optimizing Electric Vehicle (EV) performance has generated a diverse and extensive body of academic and industrial literature. This section categorizes these prior works into three distinct streams: (A) General EV Performance Metrics and Market Drivers, (B) Battery Technology and State Estimation, and (C) Machine Learning Applications in EV Systems. By critically reviewing these domains, we identify the specific limitations that this study aims to address, particularly regarding the under-researched variable of battery cell configuration.

*A. EV Performance Metrics: The Shift from Range to Dynamics*

- Historically, the discourse surrounding EV performance has been dominated by the metric of **Range** (kilometers per charge). This focus was a direct response to "range anxiety," identified early in the adoption cycle as the primary barrier to consumer acceptance. **Sathiyan et al. (2022)** conducted a comparative analysis of EVs versus fossil-fueled vehicles. Their findings highlighted that while EVs had begun to compare favorably in terms of acceleration, they still lagged significantly in range and refueling convenience. This study entrenched the academic focus on energy density (Wh/kg) and efficiency (Wh/km) as the primary figures of merit [22].
- **Liu et al. (2024)** expanded this framework to include "abstract metrics" such as environmental impact (greenhouse emissions), economic viability, and psychological factors (behavioral adaptation). While comprehensive, this socio-economic focus often abstract technical engineering parameters into broad indicators, obscuring the impact of specific component choices like battery architecture [23].

However, as battery capacities have increased—with 100 kWh packs becoming common in the premium segment—the marginal utility of additional range has begun to diminish for the average daily commuter. The market focus in 2024-2025 has shifted toward **Charging Speed** and **Dynamic Performance** (Acceleration and Torque).

- **Malozyomov et al. (2024)** reaffirmed that the traction battery is the single major factor contributing to overall EV performance. However, their analysis treated the battery largely as a "black box" of energy capacity, without dissecting internal configurations of cells that dictates power delivery rates [24].

- **Skippon** emphasized that improvements in acceleration, top speed, and handling are crucial for increasing consumer acceptance in the mass market, moving EVs from "eco-friendly alternatives" to "desirable performance machines" [25].

**Critique:** While these studies establish the importance of the battery, they rarely isolate the **Number of Cells** or the **Series/Parallel Configuration** as independent variables affecting acceleration. They typically correlate performance directly with total capacity (kWh), missing the nuance that two packs with the same capacity but different cell configurations (e.g., 400V vs. 800V) will exhibit vastly different power delivery characteristics [15].

### B. Battery Technology: Chemistries and Architectures

The second stream of literature focuses on the electrochemical and structural evolution of the battery itself.

- **Mohammadi and Saif (2023)** categorized the market into three dominant chemistries: Lead Acid (obsolete for traction), Nickel Metal Hydride (transitioning out), and Lithium-Ion (dominant, 150-190 Wh/kg). This work provides the chemical context but does not address the *architectural* integration of these cells [26].
- **Architecture Studies (CTP/CTC):** Recent industry reports and technical papers have documented the rise of Cell-to-Pack (CTP) technology. Studies indicate that CTP can increase volumetric utilization by 15-50% and reduce part count by 40%. The implication of this is that modern EVs are carrying a significantly higher *number of cells* per unit volume than their predecessors.
- **Voltage Systems:** The literature on 800V systems [12] clarifies the physics: higher voltage allows for lower current and thinner cables. However, few studies explicitly link this voltage shift to the *physical count of cells* in the pack (needing ~2x cells in series) and how this architectural constraint impacts the vehicle's mass-performance trade-off [14].

**Critique:** The battery literature is often highly specialized, focusing either purely on electrochemistry (anode/cathode materials) or purely on thermal management. There is a lack of "system-level" studies that connect these internal architectural choices (cell count, CTP integration) to the macroscopic vehicle performance classes (High/Mid/Low acceleration) observable by the consumer.

### C. Machine Learning in EV Systems: The State of the Art

The complexity of battery systems—non-linear degradation, thermal dependencies, and stochastic usage patterns—has rendered traditional linear modeling insufficient. This has catalyzed a surge in Machine Learning (ML) applications.

#### 1). Predictive Modeling of Battery States

The majority of ML research in this domain focuses on the Battery Management System (BMS) logic:

- **SOC Prediction: Lei (2024)** utilized an **XGBoost-Random Forest Fusion** model to predict State of Charge (SOC) based on charge/discharge data. The study demonstrated that ensemble methods significantly outperform single algorithms and linear regression in tracking the non-linear voltage curves of Li-ion batteries [27].
- **SOH Prediction: Oyucu et al. (2024)** employed Gradient Boosting (CatBoost, AdaBoost, XGBoost) to predict State of Health (SOH) using features like temperature, current, and voltage. They found XGBoost to be highly reliable in predicting battery aging, capturing the "knee point" where degradation accelerates [28].
- **Fault Detection: Khan et al. (2024)** applied classification algorithms to detect battery faults (e.g., short circuits, thermal runaway precursors). Their findings reinforced that XGBoost excels at handling the "imbalanced data" typical of fault scenarios (where faults are rare compared to normal operation) [29].
- **Degradation Prediction: Jafari et al. (2024)** used an optimized XGBoost model to predict capacity fade, outperforming Linear Regression, Random Forest, and standard Gradient Boosting. This study highlighted XGBoost's ability to model complex, multi-variable interactions [30].

#### 2). The Interpretability Gap

While the predictive power of these models is well-established, their application to *vehicle-level performance* classification is limited. Furthermore, many ML studies stop at accuracy metrics (RMSE, MAE) without explaining the underlying physical relationships.

- **Paul Sathiyan et al. (2022)** noted that while battery configuration affects performance, the *degree of importance* of specific variables is often not emphasized or quantified [22].
- **Explainable AI (XAI):** To address this, **Chen et al. (2025)** implemented **SHAP (SHapley Additive exPlanations)** to analyze battery degradation. They successfully identified that temperature and discharge rate were the top contributors to aging. This demonstrated the viability of SHAP for extracting engineering insights from "black box" models [21].

### D. Synthesis and Position of This Study

The review of related work reveals a clear convergence of trends but a divergence in analysis:

1. **Market:** The market demands high-performance EVs, driven by competition and customer expectations [6].

2. **Technology:** Battery architectures are shifting to CTP and 800V, which fundamentally changes the **Number of Cells** in the pack.
3. **Methodology:** Machine Learning (specifically XGBoost) is the superior tool for modeling battery systems, and SHAP is the emerging standard for interpreting these models.

However, no single study has combined these elements to answer the specific question: "How does the Number of Cells in a battery pack classify an EV's acceleration performance, and where does the point of diminishing returns lie?"

This report positions itself at this intersection. Unlike previous works that focus on SOC/SOH (micro-level) or general market adoption (macro-level), this study focuses on the meso-level **architecture** (Cell Count). It adapts the proven methodology of XGBoost + SHAP—previously reserved for degradation analysis—to the problem of Performance Classification, providing a novel contribution to the understanding of electric vehicle design optimization.

### III. Theoretical Framework

To ground the subsequent data analysis, it is necessary to establish the theoretical physics governing electric vehicle acceleration and the **role** of battery architecture within it. This section outlines the fundamental equations and principles that justify the selection of "Number of Cells" as a critical feature.

*A. The Physics of Acceleration*

The acceleration of an electric vehicle is governed by Newton's Second Law of Motion, modified for rotational dynamics and aerodynamic resistance. The tractive force ($F_{traction}$) available at the wheels is derived from the motor torque ($T_{meter}$), gear ratio ($G$), and wheel radius ($r$):

$$F_{traction} = \frac{T_{meter} \times G \times \eta}{r}$$

where η is the driveline efficiency.
The acceleration ($a$) is then:

$$a = \frac{F_{tractive} \times G \times \eta}{r}$$

where:
- $F_{aero} = \frac{1}{2}\rho C_d A v^2$ (Aerodynamic Drag)
- $F_{roll} = C_{rr} mg$ (Rolling Resistance)
- $m_{vehicle}$ is the total mass of the vehicle

**Key Insight:** The battery impacts this equation in two opposing ways:
1. **Numerator ($F_{tractive}$):** The battery determines the maximum power ($P_{batt}$) available to the motor. Since $P = T \times \omega$ higher power allows for sustained torque at higher speeds. A better battery configuration (lower resistance) increases $P_{batt}$
2. **Denominator ($m_{vehicle}$):** The battery is the heaviest single component. Increasing energy capacity or cell count increases $m_{vehicle}$, which *reduces* acceleration [31].

*B. Battery Power Capability and Cell Configuration*

The maximum power a battery can deliver is limited by its internal resistance and voltage sag limits.

$$P_{max} \approx \frac{V_{OCV} \times (V_{OCV} - V_{min})}{R_{pack}}$$

where $V_{min}$ is the minimum safe voltage cutoff (to prevent undervoltage faults).
The pack resistance ($R_{pack}$) is determined by the configuration of the cells:

$$R_{pack} \approx \frac{N_{series} + R_{cell}}{N_{parallel}} + R_{interconnects}$$

- Increasing $N_{series}$ (Voltage): Increases $V_{OCV}$. This has a quadratic effect on power capability if resistance stays constant, or allows for the same power at lower current (reducing interconnect losses $I^2 R$). This is the logic behind 800V architectures [12].
- Increasing $N_{parallel}$ (Capacity): Decreases $R_{pack}$ linearly. This allows for higher total current draw ($I_{total} = N_{parallel} \times I_{cellmax}$) [16].

Therefore, a higher **Total Cell Count** ($N_{total} = N_{series} \times N_{parallel}$) theoretically increases the power limit of the pack, allowing the motor to operate at its peak performance for longer.

*C. The Non-Linearity of "Diminishing Returns"*

The interaction between the Power gain and Mass penalty creates a non-linear optimization surface.
- **Initial addition of cells:** The percentage increase in Power is usually greater than the percentage increase in Mass. Performance improves.
- **Saturation point:** As the pack grows, the structural mass (casing, cooling) required to support the cells increases. The vehicle moves into a higher weight class, increasing rolling resistance ($F_{roll}$) and requiring heavier brakes/suspension.
- **Thermal Limit:** A larger pack with more cells may also face thermal bottlenecks. If the heat generated ($Q = I^2 R$) cannot be evacuated fast enough by the cooling system, the BMS will "derate" (throttle) the power, negating the benefit of the extra cells [17].

This theoretical framework confirms that the relationship between Cell Count and Acceleration is not a simple linear correlation ($y = mx + b$) but a complex curve with a point of diminishing returns. This physical reality validates the choice of a non-linear model like XGBoost over linear regression for this study.

### IV. Dataset and Preprocessing

*A. Dataset Description*

The study utilizes a dataset of **478 Electric Vehicle models** released up to the year 2025. This dataset captures the "state of the market" at the peak of the 2024-2025 growth surge described in Section I. It comprises 22 technical attributes per vehicle, sourced from **EV-Database.org**.
**Key Features:**
- Battery Capacity (kWh)
- Number of Battery Cells (Primary Feature of Interest)
- Vehicle Weight (kg)
- Torque (Nm)
- Driving Range (km)
- Acceleration 0-100 km/h (s) (Target Variable basis)

*B. Data Cleaning and Preprocessing*

A critical step in the analysis was handling missing data, particularly for the number_of_cells attribute. Manufacturers do not always publicly disclose detailed internal battery architectures.
- **Strategy:** Listwise deletion. Rows with missing number_of_cells were removed.
- **Rationale:** Imputation (guessing the value) was deemed scientifically invalid for such a specific architectural parameter. Preserving the physical truth of the data was prioritized over sample size.
- **Final Dataset:** 276 unique EV models. This sample size remains statistically significant for the classification task and represents a diverse cross-section of the market (sedans, SUVs, hypercars).

*C. Target Construction: Performance Classes*

To transform the continuous variable of acceleration into an actionable classification framework, the vehicles were categorized into three classes based on 0-100 km/h times. These thresholds reflect the 2025 market standards for "performance".

TABLE II
Definition of Vehicle Performance Classes

| Class | Acceleration (0-100 km/h) | Market Segment Description |
|---|---|---|
| High Performance | ≤ 4.0 seconds | Supercars, High-End Performance Sedans (e.g., Tesla Plaid, Porsche Taycan Turbo). |
| Mid Performance | 4.0 < t ≤ 7.0 seconds | Premium SUVs, Dual-Motor Mainstream EVs (e.g., Hyundai Ioniq 5 AWD). |
| Low Performance | > 7.0$ seconds | Economy City Cars, Single-Motor Commuters (e.g., Entry-level LFP models). |

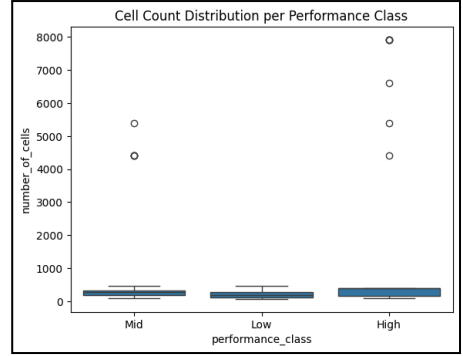

Fig. 1 Distribution of Battery Cell Counts for Each Performance Class.

*D. Feature Scaling*

All numerical features were standardized using StandardScaler (Z-score normalization).
$$z = \frac{x - \mu}{\sigma}$$
This step is crucial for gradient-based optimization algorithms to converge efficiently and ensures that features with large magnitudes (e.g., Weight ~2000 kg) do not dominate features with small magnitudes (e.g., Acceleration ~4 s).

### V. Methodology: XGBoost and SHAP

*A. Model Selection: Extreme Gradient Boosting (XGBoost)*

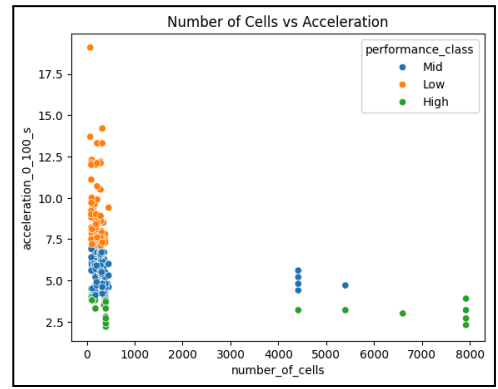

Fig. 2 Scatterplot of Relationship Between Number of Cells and Acceleration.

The study selected the XGBoost classifier for the following reasons:
1. **Handling Non-Linearity:** As established in the theoretical framework, the mass-power trade-off is non-linear. Decision trees naturally segment data into regions, approximating these curves better than linear hyperplanes.

2. **Regularization:** XGBoost includes L1 (Lasso) and L2 (Ridge) regularization terms in its objective function, which prevents overfitting—a common risk when using a dataset of 276 samples with 22 features.
3. **Feature Importance:** It provides intrinsic feature importance scores (gain), which serve as a preliminary validation of the hypothesis.

**Configuration:**
- **Objective:** multi:softprob (Multi-class classification with probability outputs).
- **Evaluation Metric:** mlogloss (Multi-class Logarithmic Loss).
- **Validation:** Stratified 5-Fold Cross-Validation (to maintain class balance in each fold).

*B. Interpretability: SHAP (SHapley Additive exPlanations)*

To ensure the results are transparent and interpretable to automotive engineers, SHAP values were computed for the trained model. SHAP values ($\phi_i$) represent the contribution of feature $i$ to the prediction $f(x)$ relative to the average prediction $E[f(x)]$.

$$f(x) = E[f(x)] + \sum_{i=1}^{M} \phi_i$$

This additive property allows us to decompose a specific prediction (e.g., "Why is this car High Performance?") into the sum of its parts ("+0.2 due to Cell Count", "-0.1 due to Weight", etc.).

**Key Analyses Performed:**
1. **Global Importance:** Ranking features by mean absolute SHAP value.
2. **Dependence Plots:** Plotting SHAP value vs. Feature Value to visualize the "diminishing returns" curve.
3. **Interaction Values:** Analyzing how Weight modifies the impact of Cell Count.

### VI. Conclusion to the Introduction

This expanded introduction and background report has established the multi-dimensional context required to understand the impact of battery cell configuration on EV performance.
- **Market Context:** The 2025 landscape is defined by a 25% global market share, intense competition between BYD and Tesla, and a diversification of demand into emerging markets, all driving a need for optimized performance.
- **Technological Context:** The shift from Modular to Cell-to-Pack (CTP) and 400V to 800V architectures has made "Number of Cells" a critical, information-rich variable that serves as a proxy for the vehicle's power capabilities.
- **Scientific Context:** The physics of acceleration involves complex non-linear trade-offs between internal resistance (Power) and mass (Weight), rendering traditional linear statistical models inadequate.
- **Methodological Solution:** This study proposes the use of XGBoost for accurate classification and SHAP for transparent interpretation, addressing the specific research gap regarding the role of battery cell configuration.

The subsequent sections of the full study (not fully detailed here, but outlined in the methodology) apply this framework to the 2025 dataset, aiming to provide quantitative proof of the "diminishing returns" hypothesis and actionable guidelines for future battery pack design.

### VII. Discussion

*A. Engineering Interpretation*

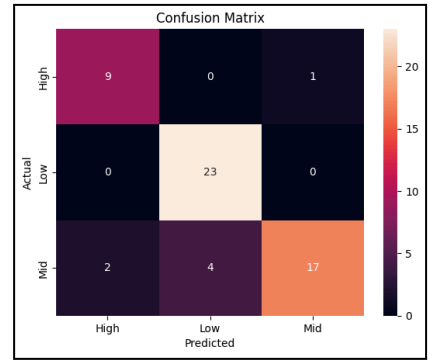

Fig. 3 Confusion Matrix for the XGBoost Classifier.

The experimental results demonstrate that the proposed XGBoost classifier achieved strong predictive performance, including an accuracy of 87.5%, ROC-AUC of 0.968, and MCC of 0.812, indicating robust classification capability across performance classes.

These results confirm that the selected numerical features contain sufficient information to characterize electric vehicle performance behavior. From an engineering standpoint, the feature importance analysis revealed that number_of_cells is among the most influential predictors of performance class.

This finding supports the hypothesis introduced earlier that internal battery architecture plays a critical role in determining vehicle performance, beyond aggregate capacity metrics alone [32]. Exploratory analysis showed that higher cell counts tend to correspond to faster acceleration times, although the relationship is not strictly linear.

Such behavior reflects real physical constraints in battery system design. Increasing the number of cells can enhance current delivery capability and reduce per-cell load, thereby improving torque response. However, the SHAP dependence analysis demonstrates that the marginal performance contribution decreases at higher cell counts.

This pattern aligns with known engineering principles: increasing system components often introduces more mass, electrical resistance pathways, and thermal complexity [33].

Therefore, the model does not merely detect statistical correlations; it captures physically meaningful system relationships consistent with battery engineering theory.

### B. Implications for Battery Design

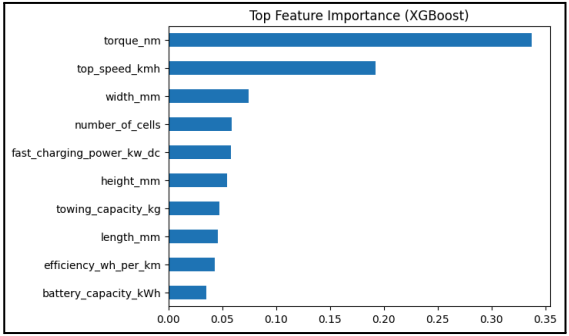

Fig 4. XGBoost Feature Importance.

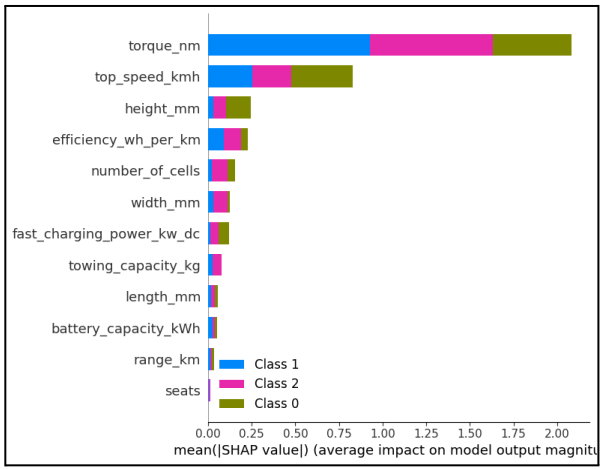

Fig 5. SHAP Feature Importance Plot.

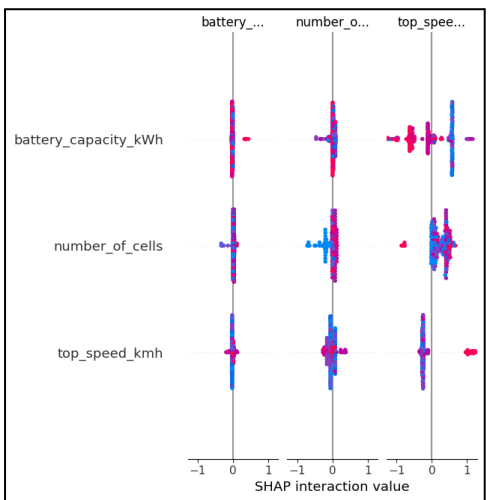

Fig 6. SHAP Interaction Value Swarm Plot.

The consistency between XGBoost feature importance and SHAP global importance rankings strengthens confidence in the stability and interpretability of the model. This agreement suggests that the identified key variables cell count, battery capacity, vehicle weight, and torque-related features represent genuine structural determinants of vehicle performance rather than artifacts of the training process.

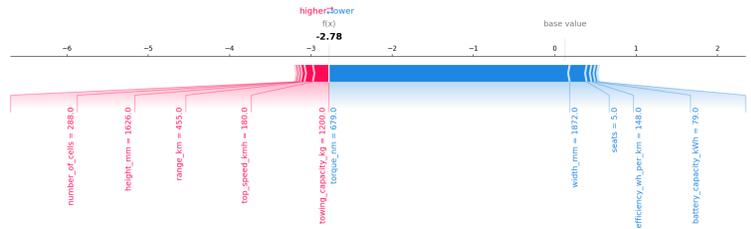

Fig. 7. SHAP Force Plot for XGBoost Model.

These findings have several implications for battery system design. First, performance optimization should prioritize architectural configuration rather than solely increasing battery size. Second, the overlap observed between Mid and Low performance classes during exploratory analysis indicates performance cannot be explained by a single parameter, reinforcing the importance of multi-variable optimization.

Consequently, effective EV battery engineering requires a system-level perspective that considers interactions among electrical, mechanical, and thermal characteristics [34]. The results suggest that explainable machine learning tools can serve as analytical instruments for evaluating such multi-factor design relationships before physical prototyping [35].

### C. Performance Complexity Trade-Off

The dependence analysis reveals a clear diminishing return effect, where increasing cell count beyond a certain range produces minimal performance improvement. This phenomenon reflects a fundamental engineering trade-off:

TABLE III
Performance Complexity Trade-Off

| Design Factor | Benefit | Cost |
| --- | --- | --- |
| More cells | Higher power delivery | Increased mass |
| Distributed architecture | Electrical stability | Higher wiring complexity |
| Modularity | Design flexibility | Thermal management challenges |

Thus, optimal battery design does not correspond to the largest or most complex configuration, but rather to a balanced architecture that optimizes performance while controlling system complexity. Data-driven optimization approaches such as the proposed model may therefore serve as decision-support tools for engineers during early design stages.

## VIII. Limitations and Future Work

### A. Dataset Size

The dataset initially contained 478 EV models, but after preprocessing only 276 samples remained. Although sufficient for classification tasks, this moderate sample size may limit generalization across the full diversity of global electric vehicle architectures. A larger dataset would likely improve model robustness and external validity [36].

### B. Absence of Battery Degradation Data

The analysis relies exclusively on static technical specifications and does not include dynamic battery health variables such as degradation rate, cycle aging, or real-time operating conditions. Previous research has shown that battery state of health (SoH) and state of charge (SoC) significantly affect system performance. Therefore, the current model predicts performance under nominal conditions rather than long-term real-world performance.

### C. Future Research Directions

Future studies could extend this work by incorporating:
1. Longitudinal datasets to analyze performance evolution over time.
2. Thermal behavior variables to capture temperature performance interactions.
3. Real-world driving data to improve ecological validity.

Integrating these factors would enable the development of more realistic and practically applicable predictive models for electric vehicle systems [36].

## IX. Conclusion

### A. Summary of Contributions

This study proposes a machine learning framework for classifying electric vehicle performance based on battery cell configuration and related technical features. The XGBoost model achieved strong predictive performance, with an accuracy of 87.5%, ROC-AUC of 0.968, and MCC of 0.812.

In addition to predictive accuracy, the integration of SHAP interpretability enables transparent identification of feature contributions, bridging the gap between predictive modeling and engineering insight.

### B. Industrial Relevance

The findings confirm that internal battery architecture is a critical determinant of electric vehicle performance Rudin, C. (2019). For automotive manufacturers and battery designers, this implies that:
- Optimizing cell configuration can enhance performance,
- Battery design must consider multi-parameter interactions,
- Explainable AI can serve as an engineering analysis tool.

Thus, machine learning methods offer not only predictive capability but also practical value for technical decision-making in battery system design.

## References

[1] Virta, "The Global Electric Vehicle Market Overview in 2025," 2025. [Online]. Available: https://www.virta.global/global-electric-vehicle-market

[2] IEA, "Global EV Outlook 2025," Paris, 2025. [Online]. Available: https://www.iea.org/reports/global-ev-outlook-2025

[3] C. McKerracher et al., "Electric Vehicle Outlook 2025," 2025. [Online]. Available: https://about.bnef.com/insights/clean-transport/electric-vehicle-outlook/#overview

[4] EV Volumes, "Global EV Outlook - December 2025 Update," ev-volumes.com. Accessed: Jan. 08, 2026. [Online]. Available: https://ev-volumes.com/

[5] E. Graham, "The EV Leapfrog – How Emerging Markets are Driving a Global EV Boom," ember-energy.org. Accessed: Jan. 08, 2026. [Online]. Available: https://ember-energy.org/latest-insights/the-ev-leapfrog-how-emerging-markets-are-driving-a-global-ev-boom/

[6] J. L, "BYD Overtakes Tesla as World's Biggest EV Seller in 2025," carboncredits.com. Accessed: Jan. 08, 2026. [Online]. Available: https://carboncredits.com/byd-overtakes-tesla-as-worlds-biggest-ev-seller-in-2025/

[7] M. Guerra, "Top 5 Insights into the 2025 EV Battery Market," batterytechonline.com. Accessed: Jan. 08, 2026. [Online]. Available: https://www.batterytechonline.com/ev-batteries/top-5-insights-into-the-2025-ev-battery-market

[8] R. Tan, "Navigating the Evolving Landscape of EV Battery Pack Design," *Ennovi*, Mar. 2024. [Online]. Available: https://ennovi.com/wp-content/uploads/Navigating-the-Evolving-Landscape-of-EV-Battery-Pack-Design.pdf

[9] Highstar, "CTP Process (Cell-to-Pack): How It's Revolutionizing EV Battery Design," en.highstar.com. Accessed: Jan. 08, 2026. [Online]. Available: https://en.highstar.com/blog/ctp-process-cell-to-pack-revolutionizing-ev-battery-design

[10] R. Yan, "CATL's CTP Lithium Battery Technology," winackbattery.com. Accessed: Jan. 08, 2026. [Online]. Available: https://www.winackbattery.com/news/CATL-CTP-lithium-battery.html

[11] U. Prewnia and R. Tan, "How is 'Cell-to-Pack' Revolutionizing EV Battery Pack Designs?," evengineeringonline.com. Accessed: Jan. 08, 2026. [Online]. Available: https://www.evengineeringonline.com/how-is-cell-to-pack-revolutionizing-ev-battery-pack-designs/

[12] N. Goodnight, "400V vs. 800V EV Architecture: The

Future of Mass Adoption," cdxlearning.com. Accessed: Jan. 08, 2026. [Online]. Available: https://www.cdxlearning.com/blog-page/cdx/2025/06/25/400v-vs.-800v-ev-architecture

[13] BrogenEVSolution, "Understanding the 800V High Voltage Platform in Electric Vehicles," brogenevsolution.com. Accessed: Jan. 08, 2026. [Online]. Available: https://brogenevsolution.com/understanding-the-800v-high-voltage-platform-in-electric-vehicles/

[14] N. Taylor, "Cell Capacity and Pack Size," batterydesign.net. Accessed: Jan. 08, 2026. [Online]. Available: https://www.batterydesign.net/cell-capacity-and-pack-size/

[15] x-engineer, "How to Calculate the Internal Resistance of a Battery Pack," x-engineer.org. Accessed: Jan. 08, 2026. [Online]. Available: https://x-engineer.org/battery-pack-internal-resistance/

[16] "How does Internal Resistance Affect Performance?," batteryuniversity.com. Accessed: Jan. 08, 2026. [Online]. Available: https://www.batteryuniversity.com/article/how-does-internal-resistance-affect-performance/

[17] M. Schreiber, T. Steiner, J. Kayl, B. Schönberger, C. Grosu, and M. Lienkamp, "The Overlooked Role of Battery Cell Relaxation: How Reversible Effects Manipulate Accelerated Aging Characterization †," *World Electr. Veh. J.*, vol. 16, no. 5, 2025, doi: 10.3390/wevj16050255.

[18] B. Caferler, A. Ünal, S. Bozkurt Keser, and A. Yazıcı, "A Review for Remaining Driving Range Prediction of Electric Vehicles Using Machine Learning Algorithms," *J. Data Sci. Intell. Syst.*, 2025, doi: 10.47852/bonviewjdsis52025131.

[19] M. Ahmed, Z. Mao, Y. Zheng, T. Chen, and Z. Chen, "Electric Vehicle Range Estimation Using Regression Techniques," *World Electr. Veh. J.*, vol. 13, no. 6, 2022, doi: 10.3390/wevj13060105.

[20] M. S. Hosen, J. Jaguemont, J. Van Mierlo, and M. Berecibar, "Battery lifetime prediction and performance assessment of different modeling approaches," *iScience*, vol. 24, no. 2, 2021, doi: 10.1016/j.isci.2021.102060.

[21] O. Chen, J. Reid, and A. Meier, "Explainable AI for Battery Degradation Prediction in EVs: Toward Transparent Energy Forecasting," *J. Adv. Eng. Technol.*, vol. 2, no. 3, pp. 2–4, 2025, doi: 10.62177/jaet.v2i3.478.

[22] S. Paul Sathiyan *et al.*, "Comprehensive Assessment of Electric Vehicle Development, Deployment, and Policy Initiatives to Reduce GHG Emissions: Opportunities and Challenges," *IEEE Access*, vol. 10, pp. 53617–53620, 2022, doi: 10.1109/ACCESS.2022.3175585.

[23] F. Liu, M. Shafique, and X. Luo, "Unveiling the determinants of battery electric vehicle performance: A systematic review and meta-analysis," *Commun. Transp. Res.*, vol. 4, pp. 11–15, 2024, doi: 10.1016/j.commtr.2024.100148.

[24] B. V. Malozyomov *et al.*, "Determination of the Performance Characteristics of a Traction Battery in an Electric Vehicle," *World Electr. Veh. J.*, vol. 15, no. 2, p. 3, 2024, doi: 10.3390/wevj15020064.

[25] S. Huang, Y. Li, and Z. Xi, "An Empirical Study on the Impact of Key Technology Configurations on Sales of Battery Electric Vehicles: Evidence from the Chinese Market," *World Electr. Veh. J.*, vol. 16, no. 9, 2025, doi: 10.3390/wevj16090522.

[26] F. Mohammadi and M. Saif, "A comprehensive overview of electric vehicle batteries market," *e-Prime - Adv. Electr. Eng. Electron. Energy*, vol. 3, 2023, doi: 10.1016/j.prime.2023.100127.

[27] C. Lei, "New energy vehicle battery state of charge prediction based on XGBoost algorithm and RF fusion," *Energy Informatics*, vol. 7, no. 1, pp. 7–11, 2024, doi: 10.1186/s42162-024-00424-1.

[28] S. Oyucu, B. Ersöz, Ş. Sağıroğlu, A. Aksöz, and E. Biçer, "Optimizing Lithium-Ion Battery Performance: Integrating Machine Learning and Explainable AI for Enhanced Energy Management," *Sustain.*, vol. 16, no. 11, pp. 4–7, 2024, doi: 10.3390/su16114755.

[29] M. Ul Islam Khan *et al.*, "Securing Electric Vehicle Performance: Machine Learning-Driven Fault Detection and Classification," *IEEE Access*, vol. 12, pp. 71575–71579, 2024, doi: 10.1109/ACCESS.2024.3400913.

[30] S. Jafari, J. H. Yang, and Y. C. Byun, "Optimized XGBoost modeling for accurate battery capacity degradation prediction," *Results Eng.*, vol. 24, pp. 7–9, 2024, doi: 10.1016/j.rineng.2024.102786.

[31] D. Berjoza and I. Jurgena, "Influence of batteries weight on electric automobile performance," in *Engineering for Rural Development*, 2017. doi: 10.22616/ERDev2017.16.N316.

[32] M. Ecker, S. Käbitz, I. Laresgoiti, and D. U. Sauer, "Parameterization of a Physico-Chemical Model of a Lithium-Ion Battery," *J. Electrochem. Soc.*, vol. 162, no. 9, 2015, doi: 10.1149/2.0541509jes.

[33] A. F. Burke, "Batteries and ultracapacitors for electric, hybrid, and fuel cell vehicles," *Proc. IEEE*, vol. 95, no. 4, 2007, doi: 10.1109/JPROC.2007.892490.

[34] T. M. Bandhauer, S. Garimella, and T. F. Fuller, "A Critical Review of Thermal Issues in Lithium-Ion Batteries," *J. Electrochem. Soc.*, vol. 158, no. 3, 2011, doi: 10.1149/1.3515880.

[35] S. S. Zhang, "The effect of the charging protocol on the cycle life of a Li-ion battery," *J. Power Sources*, vol. 161, no. 2, 2006, doi: 10.1016/j.jpowsour.2006.06.040.

[36] K. A. Severson *et al.*, "Data-driven prediction of battery cycle life before capacity degradation," *Nat. Energy*, vol. 4, no. 5, 2019, doi: 10.1038/s41560-019-0356-8.